\documentclass[10pt,twocolumn,letterpaper]{article}

\usepackage{cvpr}
\usepackage{times}
\usepackage{epsfig}
\usepackage{graphicx}
\usepackage{amsmath}
\usepackage{amssymb}
\usepackage{booktabs}

\usepackage{multirow}
\usepackage{array}

\usepackage[pagebackref=true,breaklinks=true,letterpaper=true,colorlinks,bookmarks=false]{hyperref}

\cvprfinalcopy 

\def\cvprPaperID{847} 

\ifcvprfinal\fi
\begin{document}

\title{End-to-End Deep Kronecker-Product Matching for Person Re-identification}

\author{
Yantao Shen$^{1}$\thanks{The first two authors contribute equally to this work.} \quad 
Tong Xiao$^{1}$\footnotemark[1] \quad  
Hongsheng Li$^{1}$\thanks{H. Li and X. Wang are the co-corresponding authors.} \quad 
Shuai Yi$^{2}$ \quad 
Xiaogang Wang$^{1}$\footnotemark[2] \\
$^{1}$ CUHK-SenseTime Joint Lab, The Chinese University of Hong Kong\\
$^{2}$ SenseTime Research\\
$^{1}${\tt\small \{ytshen, xiaotong, hsli, xgwang\}@ee.cuhk.edu.hk  }\\
$^{2}${\tt\small yishuai@sensetime.com} \\
}

\maketitle

\begin{abstract}
Person re-identification aims to robustly measure similarities between person images. The significant variation of person poses and viewing angles challenges for accurate person re-identification. The spatial layout and correspondences between query person images are vital information for tackling this problem but are ignored by most state-of-the-art methods. In this paper, we propose a novel Kronecker Product Matching module to match feature maps of different persons in an end-to-end trainable deep neural network. A novel feature soft warping scheme is designed for aligning the feature maps based on matching results, which is shown to be crucial for achieving superior accuracy. The multi-scale features based on hourglass-like networks and self residual attention are also exploited to further boost the re-identification performance. The proposed approach outperforms state-of-the-art methods on the Market-1501, CUHK03, and DukeMTMC datasets, which demonstrates the effectiveness and generalization ability of our proposed approach.

\end{abstract}

\section{Introduction}

Person re-identification (Re-ID) aims at finding a person of interest in an image gallery by comparing the query image of this person with all the other images in the gallery. It is an active research field in computer vision and has extensive applications in intelligent video surveillance, smart phone apps, and home robotics. Person Re-ID is closely related to yet harder than face verification, where human faces are usually well aligned, while person images show various poses and are captured from different viewing angles.

Recent years witnessed the success of deep learning methods for various computer vision problems. A large number of Convolutional Neural Network (CNN) based methods have been proposed for solving the problem of person Re-ID. Most state-of-the-art CNN based approaches aim at learning a highly non-linear mapping function that transforms person images into a common embedding space and ranking the gallery images according to their distances to the query image.

\begin{figure}[t]
   \centering
   \begin{tabular}{c@{\hspace{0mm}}c}
      &\includegraphics[scale=0.41]{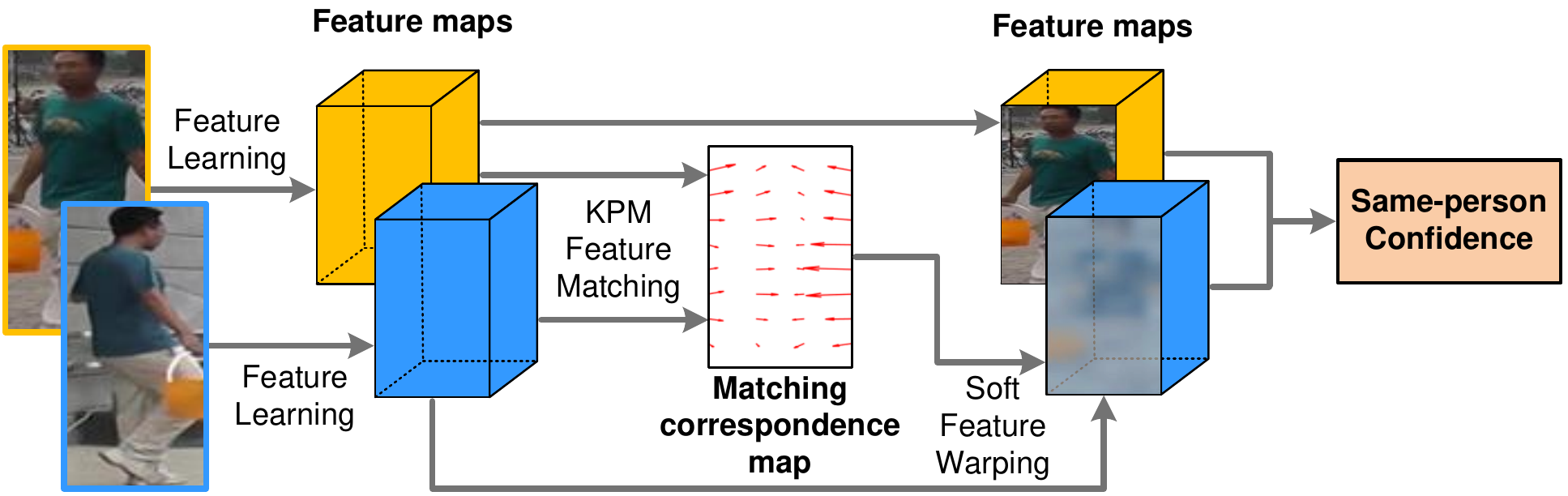}\\
   \end{tabular}
   \vspace{0pt}
   \caption{Illustration of our proposed framework with Kronecker Matching module and Soft Warping scheme for person re-identification. The KPM and soft warping are conducted between feature maps, which is end-to-end trainable in a deep network.}
   \label{fig:matching}
   \vspace{-15pt}
\end{figure}

However, most existing CNN-based methods treat each person image as an individual sample without differentiating the semantic meanings of different pixels. State-of-the-art CNN networks either utilize global average pooling (\eg, ResNet \cite{he2016deep}) or direct vectorization (\eg, VGG network \cite{Simonyan14c}) to convert the topmost feature maps into feature vectors. Directly comparing such feature vectors is ineffective because the same feature might encode different semantic concepts. The global average pooling simply averages features from all image spatial locations and abandons valuable spatial information. Features from different person regions might be compared. 
On the other hand, direct vectorization assumes that the person pose and camera viewpoints remain the same for all compared images so that features from the same image location could be compared, which is generally not true for person images from various surveillance cameras. 
In addition, existing CNN-based methods assume the same importance for all pixels in a person image. However, visual appearances of background regions are not informative and comparing them across different images leads to inaccurate similarity estimation.

To fully utilize spatial information of person images, we propose a deep neural network with Kronecker Product Matching (KPM) module to recover probabilistic correspondences between spatial regions across two images for more accurate person similarity estimation. Given the feature maps of two person images, the KPM module generates matching confidence maps to establish correspondences between them. Based on the matching confidence maps, a continuous warping scheme is adopted to deform the multi-scale feature maps of one image to match those of another image so that the feature maps of the two images could be better compared. A spatial attention mechanism is also adopted to automatically identify image regions of interest for re-identification. Extensive experiments and ablation studies on public person re-ID datasets show the effectiveness of our proposed method, which outperforms state-of-the-art approaches by large margins.

There were previous attempts on recovering person correspondences for person re-ID. Li \etal~\cite{li2014deepreid} proposed a patch matching layer that divides the pedestrian images into horizontal stripes and matches feature patches within each stripe between two images. However, the method assumes pedestrian images being vertically well aligned and cannot cope with larger deformations of persons' spatial layouts. 
More importantly, this method directly feeds matching confidence maps into classifiers to determine the person similarities. Our experiments show that such confidence maps are not discriminative enough to obtain accurate similarity estimation. In contrast, our probabilistic feature warping module generates ``softly'' warped feature maps before estimating person similarities and results in significantly improved performance. To handle unaligned images, Zheng \etal~\cite{Zheng_2015_ICCV} proposed a global matching algorithm, but the patches are divided on the image level and the features are hand-crafted. The person matching and person re-identification are performed in separate stages. In contrast, our proposed feature matching and warping modules are jointly optimized in a unified framework.


Our contribution can be summarized as threefold. First, we propose the Kronecker Product Matching module that is able to generate matching confidence maps between two images. Together with our proposed continuous warping scheme, the feature maps of person images could be stochastically deformed for end-to-end similarity learning via deep neural networks. Second, we exploit an hourglass-like network structure to generate multi-scale feature maps for person appearance encoding. The feature learning and warping are conducted at multiple scales for obtaining more robust person feature representations.
Third, we investigate a series of important factors that could significantly impact the final performance, including loss functions, input aspect ratio, and specific network design, which provides guidance for the design of the proposed and future approaches.

\begin{figure*}[t]
   \begin{center}
      \includegraphics[width=0.81\textwidth]{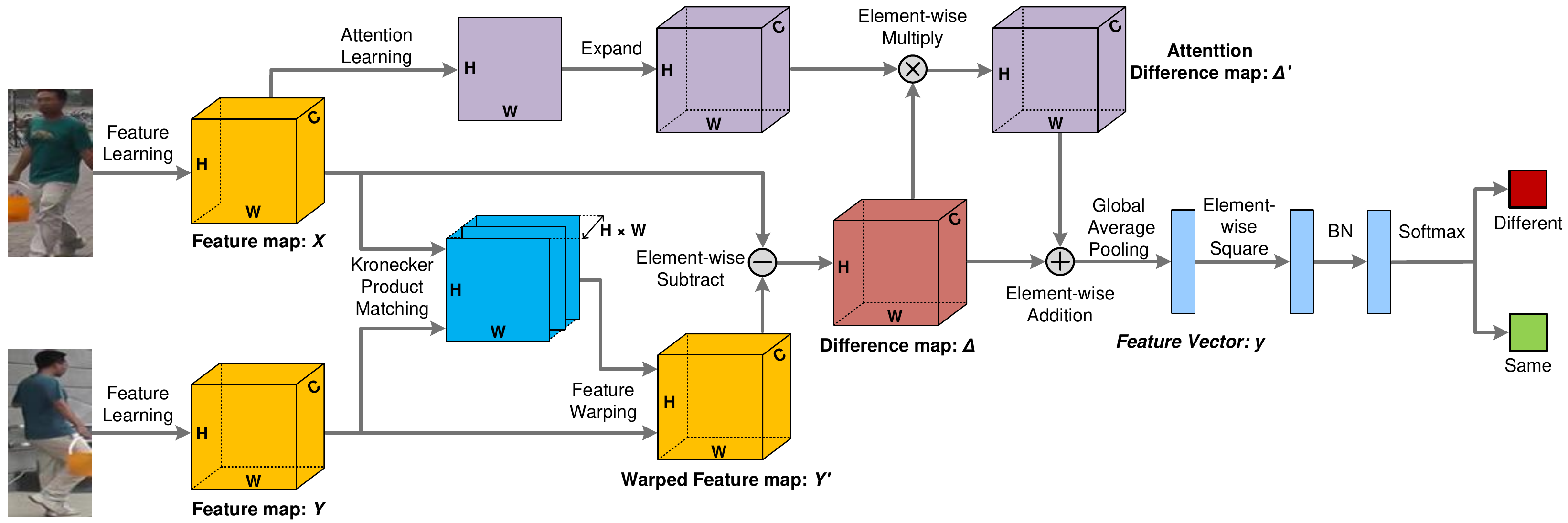}
   \end{center}{}
   \vspace{-10pt}
   \caption{Illustration of the proposed siamese-CNN network on a single scale with Kronecker Product Matching, soft feature warping, and self residual-attention modules for predicting whether the input image pair belong to the same person.}
   \label{fig:main}
   \vspace{-10pt}
\end{figure*}

\section{Related Work}
\textbf{Deep learning for person re-identification.} \ Person re-identification is a challenging problem which gains increasing attention in recent years \cite{ ahmed2015improved, cheng2016person, hamdoun2008person, wang2007shape, zhao2017person,shen2018deep,shen2017learning}. State-of-the-arts person re-identification methods adopted deep learning techniques. 
Various loss functions, including pairwise verification loss, triplet loss, and classification loss, are utilized to train the deep learning models for re-ID. Ahmed \etal \cite{ahmed2015improved} designed a Cross-Input Neighbourhood Difference CNN model for person re-identification with a pair of cropped pedestrian images as input. Li \etal \cite{li2014deepreid} proposed a patch matching layer that multiplies the activation of two images' feature maps. Both of them employed a binary verification loss function for training. Xiao \etal ~\cite{xiao2016learning,xiao2017joint} trained CNN with a classification loss to learn the deep feature of person. Ding \etal ~\cite{ding2015deep} and Cheng \etal~ \cite{cheng2016person} trained CNN with triplet samples and minimized feature distances between the same person while maximized the distances among different people. Most of these works only exploit the person identity information. Recently, pose information~\cite{zhao2017spindle,Su_2017_ICCV} and  pedestrian attribute information~\cite{Liu_2017_ICCV,lin2017improving} are also incorporated into person Re-ID systems, which enhance the performance of the trained models. A large number of metric learning methods for person re-identification have also been also proposed~\cite{paisitkriangkrai2015learning, mcfee2010metric, koestinger2012large, weinberger2009distance, yi2014deep}.

\textbf{Spatial matching for person re-identification.}
There are some previous attempts on conducting spatial matching for person re-ID. Zheng \etal~\cite{Zheng_2015_ICCV} divided images into patches and utilized a global matching algorithm with hand-crafted features. Li \etal~\cite{li2014deepreid} proposed a patch match layer for matching the features in horizontal stripes. Zheng \etal~\cite{zheng2017pedestrian} proposed a pedestrian alignment network for simultaneously aligning pedestrians within images and learning pedestrian descriptors. Bak \etal~\cite{bkak2016person} introduced a deformable patch-based model for
accommodating pose changes and occlusions. Most of these approaches divide sample into patches at image level and cannot be trained in an end-to-end manner.

\textbf{Multi-scale features for person re-identification}
Multi-scale features are beneficial for many computer vision tasks~\cite{newell2016stacked,pont2017multiscale,Zhou_2017_ICCV,wu2016model,wu20173d}. There are a few attempts of exploiting multi-scale features for person Re-ID. Liu \etal~\cite{liu2016multi} first introduced the multi-scale features into CNN training for person Re-ID, which directly downsamples image into different scales and fed them into shallow sub-networks. The outputs of these sub-networks are fused with features of the original scale for learning the similarity metric. Chen \etal~\cite{Chen_2017_ICCV_Workshops} proposed a network consisting of multiple branches for learning multi-scale features and one feature fusion branch. Different from~\cite{liu2016multi}, it incorporated person ID information for discriminative feature learning. These two works need multiple input images from multiple scales, which results in expensive computation. Qian \etal~\cite{qian2017multi} proposed a multi-scale stream layer, which was inspired by GoogLeNet~\cite{szegedy2015going} for learning features of different scales. But it needs multiple data streams for capturing multi-scale information, which results in redundant parameters. Our hourglass-like network could generate multi-scale features with fewer parameters, which is more efficiently.

\section{Approach}

Given a pair of pedestrian images, the similarity score between them is needed for determining whether these two images belong to the same person. Our proposed
approach takes two pedestrian images as inputs and outputs their similarity score with a siamese convolutional neural network, which is shown in Figure \ref{fig:main}. In this siamese-CNN, we propose Kronecker Product Matching module which will be introduced in Section \ref{sec:kronecker_product_matching} to generate matching confidence map and conduct soft warping between samples' feature maps. Please be noted that Figure \ref{fig:main} only illustrates our proposed approach in one scale. In practice, we conduct our proposed operation under different scales and combine multi-scale features together for final prediction. The multi-scale architecture we utilize is an hourglass-like structure to be introduced in Section \ref{sec:hourglass_net_with_multi_scale_kpm}.

\subsection{Kronecker Product Matching} 
\label{sec:kronecker_product_matching}

The \emph{Kronecker Product Matching} (\emph{KPM}) module is designed to spatially match two sets of convolutional feature maps. Conventional approach would convert each set to a feature vector individually with global pooling or vectorization, then followed by a fully connected layer to compute their similarity. This ignores the spatial layout between the feature maps. KPM resolves this problem by establishing correspondence for each pixel across the two sets of feature maps. Based on the matching confidence map, probabilistic warping module deforms one of them to align it with the other, and computes their differences accordingly. We introduce KPM starting from the mathematical formulation of the problem.

\subsubsection{Problem formulation} 
\label{sub:problem_formulation}

Given two input person images, the siamese-CNN outputs the two sets of convolutional feature maps, $X=\{x_1,x_2,\dots,x_M\}$ and $Y=\{y_1,y_2,\dots,y_M\}$, where $M$ is the number of pixels on each feature map, and $x_i,y_i\in\mathbb{R}^C$ are the $C$-channel feature vector at each pixel. The goal of KPM and feature warping is to produce a vector $\delta$ that describes the difference between $X$ and $Y$, which can be further fed into a distance metric (\eg, distance $d = \delta^T \Sigma \delta$ where $\Sigma$ is a Malahanobis covariance matrix) or a neural network to compute the similarity.

For each feature vector $x_i$, we would like to find the corresponding feature vector from $Y$ that matches $x_i$. Let $m_i$ be an one-hot random variable out of $M$ spatial locations, where $m_{i,j}=1$ if $x_i$ matches $y_j$. It can be assumed to follow  a multinomial distribution parameterized by $\{\beta_{i,j}\}_{j=1,\dots,M}$ with
\begin{equation}
  p(m_{i,j}=1|x_i,Y) = \beta_{i,j}.
\end{equation}
The feature vector $\hat{y}_i$ corresponding to $x_i$ could be stochastically selected as,
\begin{equation}
  \hat{y}_i = \sum_{j=1}^M m_{i,j} y_j.
\end{equation}
The difference vector $\delta_i$ at location $i$ can be computed by subtracting $x_i$ by the expected $\hat{y}_i$,
\begin{equation} \label{eq:delta-i}
  \delta_i = x_i - \mathrm{E}_{p(m_i|x_i,Y)}\left[\hat{y}_i\right] = x_i - \sum_{j=1}^M \beta_{i,j} y_j.
\end{equation}

On the other hand, different pixels could have different importance when comparing two sets of feature maps. For example, background regions in $X$ should make little contribution no matter if they have correspondences in $Y$ or not. Thus we parameterize the importance of each $x_i$ by $\alpha_i\in\mathbb{R}$ and compute the final difference vector $\delta$ between two aligned feature maps as
\begin{equation} \label{eq:delta}
  \delta = \sum_{i=1}^M \alpha_i \delta_i = \sum_{i=1}^M \alpha_i \left(x_i - \sum_{j=1}^M \beta_{i,j} y_j\right),
\end{equation}

\begin{figure}[t]
   \begin{center}
      \includegraphics[scale=0.4]{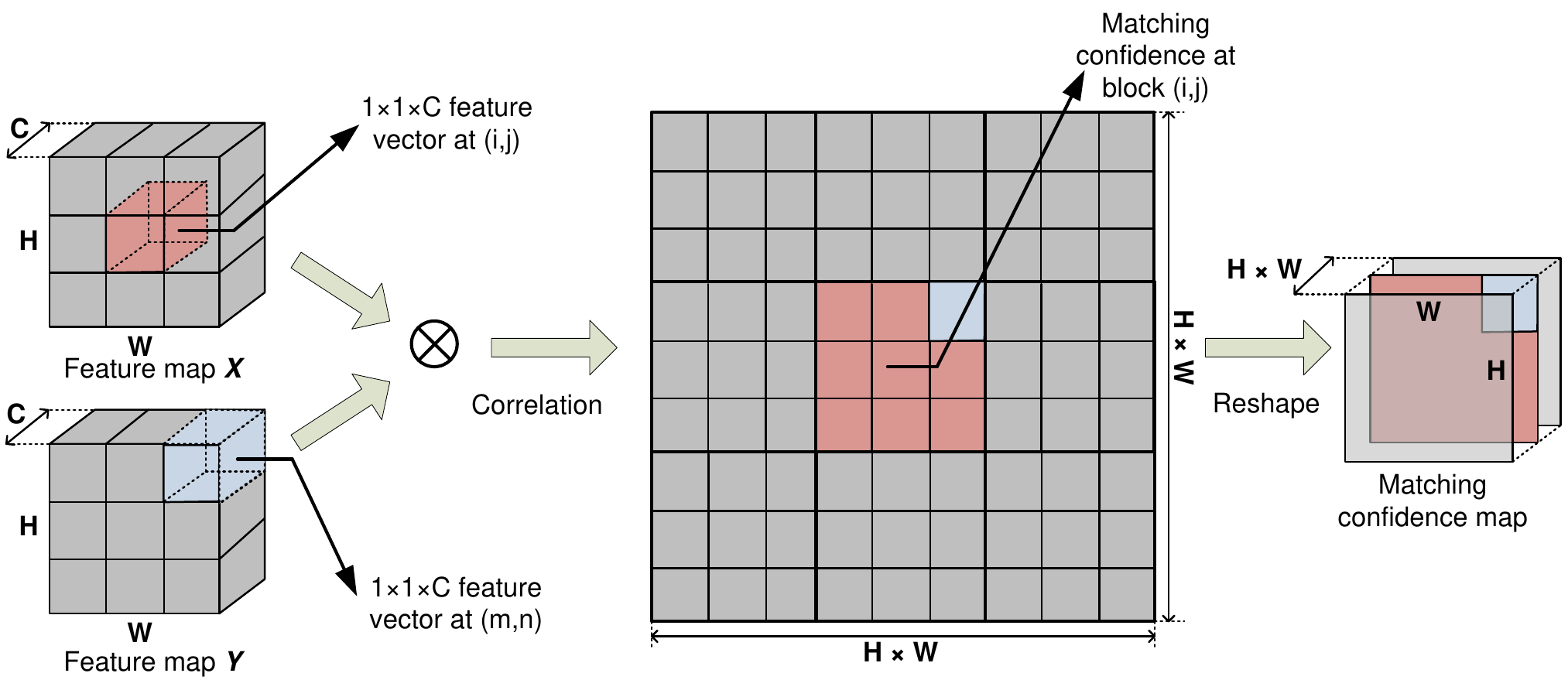}
   \end{center}
   \vspace{-10pt}
   \caption{Illustration of Kronecker Product for computing dot-product similarities between features on the two feature maps.}
   \label{fig:kron_illu}
   \vspace{-10pt}
\end{figure}

\subsubsection{Correspondence matching} 
\label{sub:correspondence_matching}
KPM computes the correspondence probability $\beta_{i,j}$ by calculating the inner product between every pair of $x_i$ and $y_j$, and then normalizing across all the spatial locations in $Y$ with the softmax function,
\begin{equation} \label{eq:beta}
  \beta_{i,j} = \frac{\exp(x_i^T y_j / \tau_{_{\text{KPM}}})}{\sum_{k=1}^M \exp(x_i^T y_k / \tau_{_{\text{KPM}}})},
\end{equation}
where $\tau_{_{\text{KPM}}}$ is the temperature hyperparameter. Lower temperature leads to lower entropy, which makes the distribution concentrate on a few high confidence locations. 

\noindent\textbf{CNN operations.} Back to the CNN representations where feature maps have height $H$ and width $W$, the feature vector at each location is indexed by row and column, \eg, $x(i,j)$ and $y(p,q)$. The above inner products between 3-D feature maps of person images are actually a straightforward extension of Kronecker Product between two 2-D matrices, as demonstrated by blue rectangles in Figure \ref{fig:main} and Figure \ref{fig:kron_illu}. The Kronecker Product results in an $H\times W\times H\times W$ tensor $K=X\otimes Y$ with
\begin{equation}
  K(i,j,p,q) = x(i,j)^T y(p,q).
\end{equation}
Each $H\times W$ submatrix can be viewed as the result of convolving an $1\times C\times 1\times 1$ filter $x(i,j)$ with the $1\times C\times H\times W$ feature maps $Y$. The whole Kronecker Product can be efficiently computed by first permuting the dimension of $X$ to $(H\times W) \times C \times 1 \times 1$, and then using it as weights to convolve with $Y$.

Following Eq.~\eqref{eq:beta}, spatial softmax normalization is applied within each submatrix to obtain the matching confidence maps
\begin{equation}
  \tilde{K}(i,j) = \text{softmax}(K(i,j,:,:)/\tau_{_{\text{KPM}}}).
  \label{eq:KPMsoftmax}
\end{equation}

After the softmax normalization we can compute the expectation in Eq.~\eqref{eq:delta-i} by matrix multiplication between $Y\in \mathbb{R}^{C\times(H\times W)}$ and the transposed $\tilde{K}^T\in\mathbb{R}^{(H\times W)\times(H\times W)}$ (where the dimensions for $Y$ are in front of those for $X$). This can be viewed as a continuous or soft warping that deforms the feature maps $Y$ to pixelwisely match the feature maps $X$. After the warping, the feature vectors across $X$ and $Y \tilde{K}^T$ are well aligned. Elementwise subtraction can be performed to obtain the difference vectors at all locations (as shown by the red cube in Figure \ref{fig:main}), $\Delta = \{\delta(i,j)\}_{i=1,\dots,H, j=1,\dots,W}$ as
\begin{equation} \label{eq:diff-map}
  \Delta = X - Y\tilde{K}^T.
\end{equation}

\subsubsection{Residual self-attention for identifying discriminative regions} 
\label{sub:residual_self_attention}

We propose to use a residual self-attention layer~\cite{wang2017residual} (purple squares in Figure \ref{fig:main}) to obtain the importance weights $\{\alpha\}$ in Eq.~\eqref{eq:delta} for $X$. 
The feature maps $X$ are fed into two consecutive sets of $1\times 1\times 256$ convolution, Batch Normalization (BN)~\cite{ioffe2015batch}, and ReLU layers. The output attention map $a\in\mathbb{R}^{1\times H\times W}$ is then spatially normalized with the softmax function
\begin{equation}
  \tilde{a}(i,j) = \frac{\exp(a(i,j) / \tau_{_{\text{RSA}}})}{\sum_{i,j} \exp(a(i,j) / \tau_{_{\text{RSA}}})},
  \label{eq:selfattsoftmax}
\end{equation}
with the temperature hyperparameter $\tau_{_{\text{KPM}}}$. We simply set $\alpha = 1 + \tilde{a}$, which means the image regions related to distinct regions should be strengthened by greater-than-1 weights.


At last, the above map of difference vectors in Eq.~\eqref{eq:diff-map} is elementwisely weighted by $1 + \tilde{a}(i,j)$ and summed up to generate the final difference vector for classification,
\begin{equation}
  \tilde{\delta} = \sum_{i,j} (1+\tilde{a}(i,j)) \delta(i,j).
\end{equation}
An elementwise-square operation followed by a BN and a fully-connected layer is adopted to output the final confidence on whether the input pair belongs to a same person.

\begin{figure*}[t]
   \begin{center}
      \includegraphics[width=0.81\textwidth]{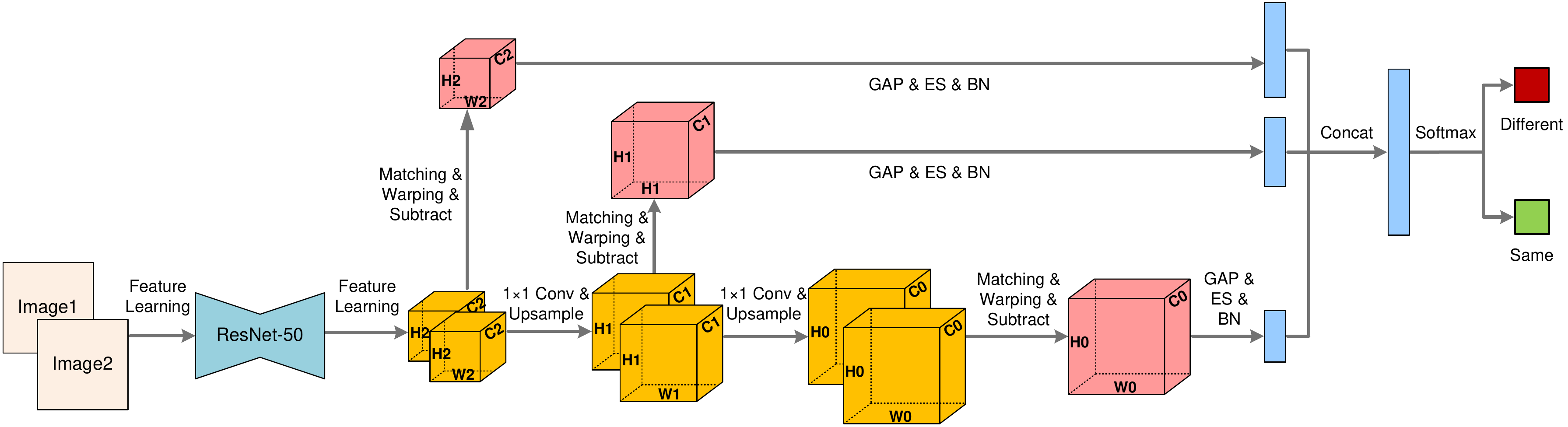}
   \end{center}
   \vspace{-10pt}
   \caption{Illustration of the hourglass network for generating multi-scale feature maps with Kronecker Product Matching. (``GAP'' denotes global average pooling and ``ES'' denotes elementwise square.)}
   \label{fig:hourglass}
   \vspace{-15pt}
\end{figure*}

\subsubsection{Discussion with the existing approach} 
\label{sub:discussion}

The patch matching layer in \cite{li2014deepreid} considers spatial matching between convolutional feature maps for person Re-ID. However, they discard the original feature maps and utilize only the matching results to determine the similarity. We tried such a variant based on our framework, but the result shows that directly using matching confidence maps for similarity estimation is inferior to our proposed approach with not only feature matching but also feature warping operations. In addition, the correspondence matching in \cite{li2014deepreid} is only conducted within each horizontal stripe, rather than the whole feature maps, which further restricts its discriminative ability on two vertically unaligned images.

\subsection{Hourglass network for multi-scale KPM} 
\label{sec:hourglass_net_with_multi_scale_kpm}

Based on the KPM and residual self-attention modules, for fully exploiting the multi-scale information, we adopt a hourglass-like structure \cite{newell2016stacked} to generate multi-scale feature maps. An overview of our hourglass network is illustrated in Figure \ref{fig:hourglass}. In our network, input images are resized to $256 \times 128$. The network generates feature maps from three different scales for feature matching, warping, and similarity estimation. The heights and widths of these feature maps are $8 \times 4$ (scale-1), $16 \times 8$ (scale-2) and $32 \times 16$ (scale-3), respectively. Following the structure in \cite{newell2016stacked}, to gradually generate feature maps with larger sizes, we conduct $1 \times 1$ convolution on lower-resolution feature maps to halve the channel dimension followed by bilinear upsampling to double the height and width dimension. The previous feature maps with the same spatial sizes are also processed by a $1\times 1$ convolution and elementwisely added to the output multi-feature maps. For feature maps of each scale, we conduct KPM, soft feature warping, elmentwise subtraction with residual self-attention, and the following operations in Section \ref{sec:kronecker_product_matching} to obtain the feature vector for classification. The feature vectors from the three scales are concatenated and fed in to a fully connected layer for estimating the final similarity score.

\section{Experiments}

To evaluate the overall performance and individual component of the proposed approach on person Re-ID, we conduct experiments on three popular public datasets.

\subsection{Datasets and evaluation metric}
\label{dataset}


\textbf{CUHK03}~\cite{li2014deepreid} has 14,097 images of 1,467 identities captured by two cameras from the CUHK campus. It contains  manually annotated images and automatically detected images. We use manually annotated ones in this paper.

\textbf{Market-1501}~\cite{zheng2015scalable} is a large-scale person Re-ID dataset which has 12,936 images for training and 19,732 images for testing. There are totally 1,501 identities in this dataset, captured from a real city market. All the images in this dataset are obtained by the DPM detector~\cite{felzenszwalb2010object}. 

\textbf{DukeMTMC}~\cite{ristani2016MTMC} is also collected from a campus. It contains 1,812 identities from eight cameras. Among these 1,812 identities, 1,404 identities appear in more than two cameras. Here, we follow the setup in ~\cite{zheng2017unlabeled} to divide DukeMTMC dataset into two parts: 16,522 images with 702 identities for training and 19,989 images with 702 identities for testing.  

We adopt mean average precision (mAP) and top-1, top-5, top-10 cumulative matching characteristics (CMC) accuracies as evaluation metric. For different datasets,  different CMC computation methods are adopted following each dataset's own convention to calculate the final performance.

\renewcommand{\multirowsetup}{\centering}
\begin{table*}
   \small
   \begin{center}
      \begin{tabular}{ccccccccc}
         \toprule
         \multirow{2}{*}{Factors}&
         \multirow{2}{*}{Designs}&
         \multicolumn{2}{c}{Market-1501~\cite{zheng2015scalable}}&
         \multicolumn{2}{c}{CUHK03~\cite{li2014deepreid}} &
         \multicolumn{2}{c}{DukeMTMC~\cite{ristani2016MTMC}}\\
         & &mAP&top-1&mAP&top-1&mAP&top-1\\
         \midrule
         \multirow{4}{*}{Loss functions} & Cross-entropy loss~\cite{xiao2017joint} &59.8 &81.4 &62.7&70.8 &40.7&62.5\\
         & OIM loss~\cite{xiao2017joint} &60.9 & 82.1 &72.5&77.5& 47.7 & 68.1\\
         & Triplet loss~\cite{hermans2017defense}  & 67.9 & 85.1 &80.7&84.3&54.6&73.1\\
         & Difference verification loss*  & \textbf{68.8} &\textbf{86.4} & \textbf{82.9} & \textbf{85.2} &\textbf{55.5}& \textbf{75.3}\\ \hline
         \multirow{4}{*}{Input aspect ratio}& 1:1 (ImageNet pretrained) &66.6 &86.1 &79.8&\textbf{85.4}&54.9&57.1\\
         & 1:1 (random initialization) &30.2&58.9&60.8&66.0&23.2&45.6\\
         & 2:1 (random initialization) &32.4&61.3&61.5&67.6&24.9&46.6\\
         & 2:1 (ImageNet pretrained)* & \textbf{68.8} &\textbf{86.4} & \textbf{82.9} & 85.2 &\textbf{55.5}&\textbf{75.3}\\ \hline
         \multirow{3}{8em}{Feature difference\\ normalization}& Absolute value &66.&85.0&81.7 &84.1 &54.6&73.8\\
         & Square before pooling &64.3 &\textbf{86.7} &72.7&78.5&50.9&\textbf{75.4}\\
         & Square after pooling* & \textbf{68.8} &86.4 & \textbf{82.9} & \textbf{85.2} &\textbf{55.5}&75.3\\
         \bottomrule
      \end{tabular}
   \end{center}
   \vspace{0pt}
   \caption{Results by the base ResNet-50 network with different components on the Market-1501~\cite{zheng2015scalable}, CUHK03~\cite{li2014deepreid} and DukeMTMC~\cite{ristani2016MTMC} dataset. ``*'' denotes the component we use in our final model.}
   \label{tab:baseline}
   \vspace{-15pt}
\end{table*}

\subsection{Implementation details}
\label{imp}
We implement our KPM hourglass network based on the ResNet-50~\cite{he2016deep}, which is pretrained on the ImageNet dataset. All the input images are resized to $256 \times 128$. The random horizontal flip and random erasing~\cite{zhong2017random} are utilized for data augmentation. The positive-to-negative pair ratio for the each mini-batch is set to 1:3 for the Market-1501 dataset and 1:4 for the CUHK03 and DukeMTMC datasets. The temperature parameter $\tau_{_{\text{KPM}}}$ in Eq.~(\ref{eq:KPMsoftmax}) is set to 0.05 and $\tau_{_{\text{RSA}}}$ in Eq.~(\ref{eq:selfattsoftmax}) is set to 1.0. We adopt SGD optimizer for model training and the initial learning rate is set to 0.01 which gradually drops to 0.001 after 50 epochs. It is then fixed for another 50 training epochs.

\subsection{Base model design and analysis}
\label{baseline}
For the Re-ID task, we observe that some basic network designs and training strategies have great impact on the final performance. In this section, we investigate loss functions, input image aspect ratio and difference feature normalization with a base ResNet-50 network. The analysis results could validate the design and hyperparameter setting of our proposed model in Figure \ref{fig:main}.

\begin{figure}[t]
   \begin{center}
      \includegraphics[scale=0.42]{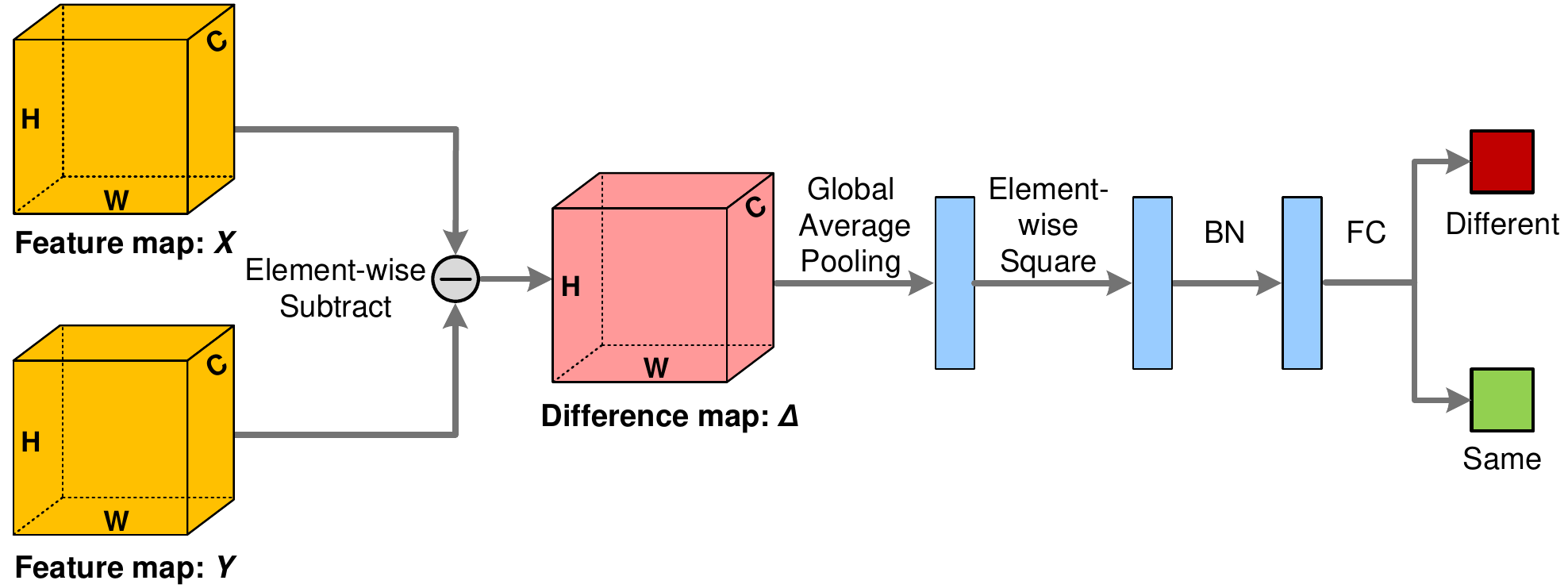}
   \end{center}
   \vspace{-10pt}
   \caption{Illustration of the base ResNet-50 network with the proposed pair difference classification}
   \label{fig:baseline}
   \vspace{-12pt}
\end{figure}

\textbf{Loss function.} To investigate the effectiveness of the different loss functions on learning person features, we train a base ResNet-50 network with the cross-entropy (softmax) loss on predicting person identities \cite{xiao2016learning}, with the triplet loss on learning feature differences between samples \cite{hermans2017defense}, and with the proposed difference verification loss on verifying the similarities of input pairs. For training with our proposed difference verification loss, we utilize a siamese-structure, which is illustrated in Figure \ref{fig:baseline}. A pair of images are fed into the siamese network. The feature difference between the two images are obtained by elementwise subtraction between the output feature maps of the last convolutional layer. The difference feature is then averaged across all locations and followed by a final fully-connected layer to compute the similarity confidence. Finally, the binary cross-entropy loss is adopted as the loss function.

For fair comparison, all the models are pretrained on ImageNet and we train the models with the cross-entropy loss~\cite{xiao2017joint}, OIM loss~\cite{xiao2017joint} and triplet loss~\cite{hermans2017defense} following their original training scheme. The results by different loss functions are reported in Table \ref{tab:baseline}. The meanAP and top-1 accuracy of these three loss functions are lower than that of the proposed difference verification loss function, which demonstrates the effectiveness of our difference verification loss function. In addition, the revised triplet loss in~\cite{hermans2017defense} needs a hard negative mining stage, which requires evaluating all training images with a partially trained model for hard negative sampling. We use the proposed siamese-ResNet-50 with difference verification loss in Figure \ref{fig:baseline} for the following experiments in this section.

\textbf{Input aspect ratio.} Conventional deep learning based methods adopt state-of-the-art deep networks (such as GoogLeNet and ResNet), for learning person features, which usually input images of squared shapes (\eg, $224\times224$). However, since most person images have an height:width aspect ratio of roughly 2:1, we argue that resizing the images to square shapes would over-stretch the images in the horizontal dimension and hinder the feature learning capability. To validate our argument on the input image aspect ratio, we run experiments with different image aspect ratios corresponding to input image sizes of $256\times128$ and $224\times224$. The proposed model is initialized with random parameters or ImageNet pretrained parameters. The results are shown in Table \ref{tab:baseline}, which shows that the network trained with input aspect ratio 2:1 outperforms those trained with 1:1 aspect ratio. The 2:1 input aspect ratio results in gains of 2.2\%, 0.3\%, 1.7\% on Market-1501, CUHK03, and DukeMTMC datasets with random initialization in terms of meanAP and achieve similar gains with ImageNet pretrained parameters.

\textbf{Features difference normalization.} For validating the importance of the elementwise square on the feature difference vector, we replace the elementwise square with absolute value operation in our base model in Figure \ref{fig:baseline}, the meanAP drops by 1.9\%, 1.2\% , 0.9\% and top-1 accuracies drop by 1.4\%, 1.1\%, 1.5\% on three datasets. Then we also test removing the elementwise square from our base model and the model fails to converge. Furthermore, we also investigate the impact of where to conduct elementwise square. When we move the elementwise square before global average pooling, the meanAP drop by 3.7\%, 7.8\% , 4.6\%  on Market-1501, CUHK03, and DukeMTMC respectively. We also remove the subtraction and elementwise square operation, and simply concatenating the feature vectors of image pair for verification, but the training cannot converge.

\begin{table}
   \small
   \setlength{\tabcolsep}{4pt}
      \begin{center}
      \begin{tabular}{lccccc}
         \toprule
         \multirow{2}{*}{Methods}&
         \multirow{2}{*}{Ref}&
         \multicolumn{4}{c}{Market-1501~\cite{zheng2015scalable}}\\
         & &mAP&top-1&top-5&top-10\\
         \midrule
         DGD~\cite{xiao2016learning}&CVPR'16 &31.9 &59.5 &~ - &~ -\\
         CADL~\cite{Lin_2017_CVPR}&CVPR'17 &47.1&73.8&~ - &~ - \\
         P2S~\cite{Zhou_2017_CVPR}&CVPR'17 &44.3&70.7&~ - &~ - \\
         MSCAN~\cite{Li_2017_CVPR}&CVPR'17 &53.1&76.3&~ - &~ - \\
         SSM~\cite{bai2017scalable}&CVPR'17 &68.8&82.2&~ - &~ -\\
         OIM Loss~\cite{xiao2017joint}&CVPR'17& 60.9 & 82.1  &~ - &~ - \\
         SpinNet~\cite{zhao2017spindle}&CVPR'17 &~ -&76.9&91.5&94.6\\
         JLML~\cite{gong2017person}&IJCAI'17 &65.5&85.1&~ - &~ -\\
         VI+LSRO~\cite{zheng2017unlabeled}&ICCV'17 &66.1&84.0&~ -&~ -\\
         OL-MANS~\cite{Zhou_2017_ICCV}&ICCV'17 &~ -&60.7&~ -&~ -\\
         PDC~\cite{Su_2017_ICCV}&ICCV'17 &63.4&84.1&92.7&94.9\\
         PA~\cite{zhao2017deeply}&ICCV'17 & 63.4 & 81.0 &92.0&94.7\\
         SVDNet~\cite{Sun_2017_ICCV}&ICCV'17  &62.1&82.3&92.3&95.2\\
         Ours & & \textbf{75.3} &\textbf{90.1} &\textbf{96.7} & \textbf{97.9}\\
         \bottomrule
      \end{tabular}
   \end{center}
   \vspace{-5pt}
   \caption{mAP, top-1, top-5, and top-10 accuracies of compared methods on the Market-1501 dataset~\cite{zheng2015scalable}.
   }
   \label{tab:market}
   \vspace{-15pt}
\end{table}

\begin{table}
\setlength{\tabcolsep}{4pt}
   \small
   \begin{center}
      \begin{tabular}{lccccc}
         \toprule
         \multirow{2}{*}{Methods}&
         \multirow{2}{*}{Ref}&
         \multicolumn{4}{c}{CUHK03~\cite{li2014deepreid}}\\
         & &mAP&top-1&top-5&top-10\\
         \midrule
         DGD~\cite{xiao2016learning}&CVPR'16&~ - &72.6 &91.6 &95.2\\
         MSCAN~\cite{Li_2017_CVPR}&CVPR'17&~ -&74.2&94.3&97.5\\ 
         SSM~\cite{bai2017scalable}&CVPR'17&~ -&76.6&94.6 & 98.0\\
         SpinNet~\cite{zhao2017spindle}&CVPR'17&~ -&88.5&97.8 & 98.6\\
         OIM Loss~\cite{xiao2017joint}&CVPR'17& 72.5 & 77.5  &~ - &~ - \\
         Quadruplet~\cite{Chen_2017_CVPR}&CVPR'17&~ -&75.5&95.2&99.2\\
         JLML~\cite{gong2017person}&IJCAI'17&~ -&83.2&98.0 & 99.4\\
         OL-MANS~\cite{Zhou_2017_ICCV}&ICCV'17&~ -&61.7&88.4&95.2\\
         PA~\cite{zhao2017deeply}&ICCV'17 &~ - & 85.4 &97.6&99.4\\
         SVDNet~\cite{Sun_2017_ICCV}&ICCV'17 &84.8& 81.8&95.2&97.2\\
         VI+LSRO~\cite{zheng2017unlabeled}&ICCV'17&87.4&84.6&97.6&98.9\\
         PDC~\cite{Su_2017_ICCV}&ICCV'17&~ -&88.7&\textbf{98.6}&\textbf{99.6}\\
         MuDeep~\cite{qian2017multi}&ICCV'17&~ - &76.3 &96.0 &98.4\\
         Ours & & \textbf{89.2} &\textbf{91.1} &98.3 & 99.1\\
         \bottomrule
      \end{tabular}
   \end{center}
   \vspace{-5pt}
   \caption{mAP, top-1, top-5 and top-10 accuracies by compared methods on the CUHK03 dataset~\cite{li2014deepreid}}
   \label{tab:cuhk}
   \vspace{-15pt}
\end{table}

\begin{table}{\hspace{-1.0cm}}
   \small
   \setlength{\tabcolsep}{3pt}
   \begin{center}
      \begin{tabular}{lccccc}
         \toprule
         \multirow{2}{*}{Methods}&
         \multirow{2}{*}{Ref}&
         \multicolumn{4}{c}{DukeMTMC~\cite{ristani2016MTMC}}\\
         & &mAP&top-1&top-5&top-10\\
         \midrule
         BoW+KISSME~\cite{zheng2015scalable}&ICCV'15 & 12.2 &25.1&~ -&~ -\\
         LOMO+XQDA~\cite{liao2015person}&CVPR'15 & 17.0 &30.8&~ -&~ -\\
         ACRN~\cite{schumann2017person}&CVPR'17 & 52.0 &72.6&84.8&88.9\\
         OIM Loss~\cite{xiao2017joint} &CVPR'17& 47.4 & 68.1  &~ - &~ - \\
         Basel+LSRO~\cite{zheng2017unlabeled}&ICCV'17 &47.1&67.7&~ -&~ -\\
         SVDNet~\cite{Sun_2017_ICCV}&ICCV'17 & 56.8 &76.7&86.4&89.9\\
         
         Ours & & \textbf{63.2} &\textbf{80.3} &\textbf{89.5} & \textbf{91.9}\\
         \bottomrule
      \end{tabular}
   \end{center}
   \vspace{-5pt}
   \caption{mAP, top-1, top-5, and top-10 accuracies by compared methods on the DukeMTMC dataset~\cite{ristani2016MTMC}}
   \label{tab:duke}
   \vspace{-15pt}
\end{table}

\subsection{Comparison with existing person Re-ID methods}
\label{compare}

\textbf{Results on Market-1501 dataset.} Table \ref{tab:market} shows the experimental results of our approach and state-of-the-art methods on the Market-1501 dataset. Our approach results in the best performance on meanAP, top-1, top-5, and top-10 CMC accuracies, which demonstrates the effectiveness of the proposed method on this dataset. 
The SSM~\cite{bai2017scalable} method estimates the similarity between image pairs in the context of other pairs of instances. Compared with the SSM approach, our framework gains 6.5\% and 7.9\% in terms of meanAP and CMC top-1. 
Part Aligned (PA)~\cite{zhao2017deeply} method jointly solves body detection and person Re-ID. Our method results in a meanAP of 75.3\% and a CMC top-1 accuracy of 90.1\%, which outperforms the PA by 63.4\% and 81.0\%. Similarly, SpindleNet (SpinNet)~\cite{zhao2017spindle} and PDC~\cite{Su_2017_ICCV} also integrate the human pose information in the person Re-ID pipeline which need human pose estimation pretrained model. Our approach does not need any pose estimation model yet still outperforms these methods. 

\textbf{Results on CUHK03 dataset.}
The Re-ID results on CUHK03 dataset is shown in Table \ref{tab:cuhk}. The meanAP and top-1 CMC accuracy of our framework are 89.2\% and 91.1\% which outperform those of state-of-the-art ones. For top-5 and top-10 CMC accuracies, PDC~\cite{Su_2017_ICCV} yields slightly better performance than ours. However, PDC needs human pose estimation pretrained model, which is not utilized in our framework. 
MuDeep~\cite{qian2017multi} learns discriminative features with different spatial scales and locations of person images. Our method improves the top-1 accuracy by 14.8\% compared with MuDeep.
Verif-Identif.+LSRO (VI+LSRO)~\cite{zheng2017unlabeled} utilizes additional training data generated by GAN. Our method does not utilize any additional training data but still outperforms it.

\textbf{Results on DukeMTMC dataset.}
In Table \ref{tab:duke}, we show the results of our framework and those by state-of-the-art ones. Our method  outperforms all the compared frameworks. The gains on meanAP and CMC top-1 accuracy by our proposed framework are 
6.4\% and 3.6\% compared with the state-of-the-art SVDNet~\cite{Sun_2017_ICCV}. ACRN~\cite{schumann2017person} also integrates person attribute information into the training process. However, our method outperforms it by 6.3\% and 7.5\% on mAP and top-1 accuracy. 

\begin{figure*}[t]
   \centering
   \begin{tabular}{c@{\hspace{0mm}}cc}
      &\includegraphics[scale=0.35]{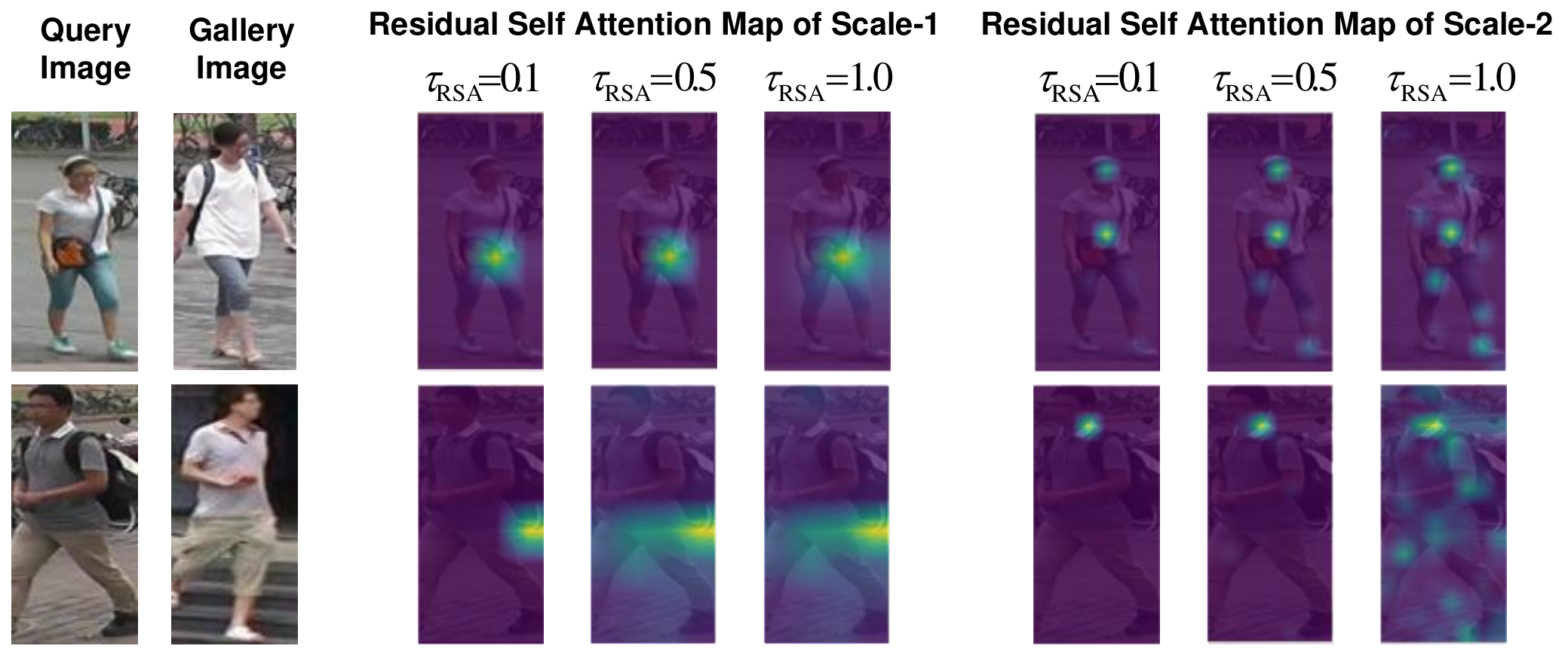}
      &\includegraphics[scale=0.35]{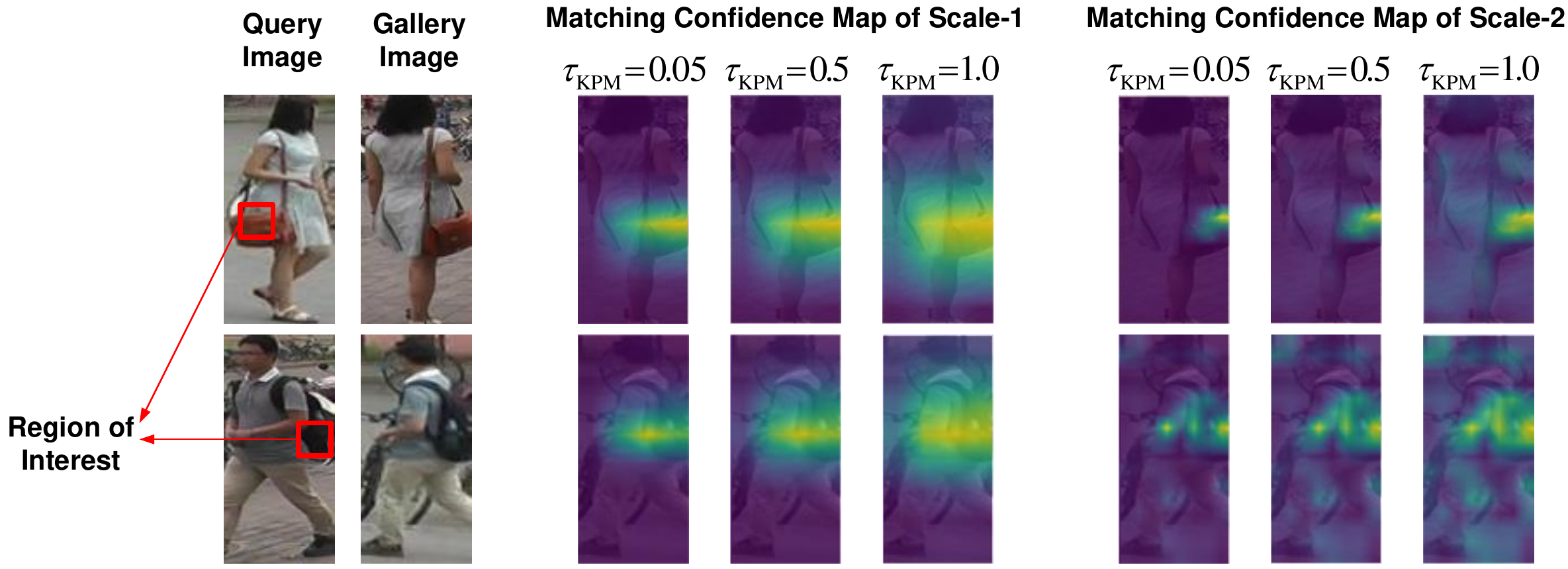}
   \end{tabular}
   \vspace{0pt}
   \caption{(Left) example attention maps by our approach of scale-1 and scale-2. Different distinct regions are focused to distinguish the pairs of person images. (Right) example matching confidence maps of feature vectors in red rectangles to those of another image.}
   \label{fig:kron_map_and_att_map}
   \vspace{-5pt}
\end{figure*}

\begin{table*}
   \small
   \begin{center}
      \begin{tabular}{lccccccc}
         \toprule
         \multirow{2}{*}{Methods}&
         \multicolumn{2}{c}{Market-1501~\cite{zheng2015scalable}}&
         \multicolumn{2}{c}{CUHK03~\cite{li2014deepreid}} &
         \multicolumn{2}{c}{DukeMTMC~\cite{ristani2016MTMC}}\\
         &mAP&top-1&mAP&top-1&mAP&top-1\\
         \midrule
         Baseline &68.8 &86.4 &82.9&85.2 &55.5&75.3\\
         Baseline+KPM &70.6 & 87.4 &85.5&88.3& 56.0 & 76.3\\
         Baseline+KPM w/o warp &62.6&86.3&71.7&80.8&50.8&75.1\\
         Baseline+KPM+RSA & 72.0 & 88.7 &86.2&89.3&58.1&77.0\\
         Baseline+HG &69.9 &87.7 &84.1&87.0&60.4&79.0\\
         Baseline+HG ML &69.1 &86.8 & -& -& -& -\\
         Baseline+KPM+RSA+HG & \textbf{75.3} &\textbf{90.1} & \textbf{89.2} & \textbf{93.4} &\textbf{63.2}&\textbf{80.3}\\
         \bottomrule
      \end{tabular}
   \end{center}
   \vspace{0pt}
   \caption{Ablation studies on the Market-1501~\cite{zheng2015scalable}, CUHK03~\cite{li2014deepreid} and DukeMTMC~\cite{ristani2016MTMC} datasets. ``+KPM'': with Kronecker Product Matching and soft warping modules. ``+KPM w/o warp'': with only Kronecker Product Matching module.  ``+RSA'': with residual self-attention. ``+HG'': with hourglass structure. ``+HG ML'' with hourglass structure and using multiple losses for different scales.}
   \label{tab:ablation}
   \vspace{-15pt}
\end{table*}

\subsection{Ablation Study}
\label{ablation}

\textbf{Importance of different components.}
In this section, we investigate the effectiveness of each component in our proposed model by conducting a series of ablation studies on all three datasets. We treat the base ResNet-50 with pair difference classification in Section \ref{baseline} as baseline model in this section. We first study the effectiveness of our Kronecker Product Matching and feature warping modules. We first train our baseline model with the KPM module, named \emph{Baseline+KPM}. Its meanAP and CMC top-1 accuracy increase by 1.8\%, 1.0\% on the Market-1501 dataset, 2.6\%, 3.1\% on the CUHK03, and 0.5\%, 1.0\% on the DukeMTMC dataset compared with baseline model, which demonstrates that conducting soft warping with KPM helps align feature maps to enhance the performance. 
For validating the effectiveness of soft warping following KPM, we remove the soft warping operation and subtraction between warped feature maps which is illustrated in Eq. (\ref{eq:diff-map}). Instead, we directly take the Kronecker Product Matching results from Eq. (\ref{eq:KPMsoftmax}) and input them into a few of convolution and fully-connected layers to obtain the similarity confidence (denoted as \emph{Baseline+KPM} w/o warp). Such an approach could be regarded as a extension of FPNN~\cite{li2014deepreid}. The mAP of this approach drops by 4.7\%, 11.2\% and 4.3\% on the datasets and top-1 accuracies drop by 0.1\%, 4.4\% and 0.2\%. 

Then we investigate the importance of the proposed residual self-attention mechanism, which weights more on representative regions for verification. (denoted as \emph{Baseline+KPM+RSA}). The meanAP has 1.4\%, 0.7\% and 2.1\% increase on the Market-1501, CUHK03, and DukeMTMC datasets compared with Baseline+KPM model and the top-1 accuracies improve Baseline+KPM by 1.3\%, 1.0\% and 0.7\% respectively. The improvements show the residual self-attention can help to implement the importance weights {$\alpha$} in Eq. (\ref{eq:delta}). Next, we investigate the importance of the hourglass structure for obtaining multi-scale feature maps. Based on the baseline model, we integrate hourglass structure (denoted as \emph{Baseline+HG}). Its mAP and top-1 accuracy increase by 1.1\%, 1.3\% on Market-1501, 1.2\%, 1.8\% on CUHK03, and 4.9\%, 3.7\% on DukeMTMC compared to the baseline model, which illustrate that integrating the multi-scale information into the framework is beneficial for distinguishing appearances of different people. 

Furthermore, to investigate different hourglass structures' impact. we also compare the results of two ways of combining information from multiple scales: 1) concatenating the feature vectors from different scales and to predict the similarity confidence (Baseline+HG); and 2) computing similarity confidences separately for each scale and averaging the confidences of multiple scales (denoted as \emph{Baseline+HG ML}). Compared with \emph{Baseline+HG}, the mAP and top-1 accuracy of Baseline+HG ML drop by 0.7\% and 0.9\% on Market-1501. Finally, the \emph{Baseline+KPM+RSA+HG} denotes our proposed overall framework. Compared with our baseline model, its mAP and top-1 accuracy have 6.5\% and 3.7\% increase on Market-1501, 6.3\% and 8.2\% increase on CUHK03, 7.7\% and 5.0\% increase on DukeMTMC. 

\textbf{Temperature hyperparameters for KPM and RSA.}
In this section, we survey the temperature hyperparameters $\tau_{_{\text{KPM}}}$ in Eq. (\ref{eq:KPMsoftmax}) for Kronecker Product Matching (KPM) and $\tau_{_{\text{RSA}}}$ in Eq. (\ref{eq:selfattsoftmax}) for Residual Self-Attention (RSA). Decreasing $\tau_{_{\text{KPM}}}$ and $\tau_{_{\text{RSA}}}$ leads to lower entropy, thus the matching confidence map and the attention map would be more concentrated. In our approach, we prefer the matching maps to concentrate on a few high confidence locations while attention maps are dispersed to capture the whole foreground human region. Different temperatures' impact on confidence maps and attention maps are shown in Figure \ref{fig:kron_map_and_att_map}. As illustrated in Table \ref{tab:temp}, such assumptions are validated by our experiments on the two temperature hyperparameters, where 0.05 is set for $\tau_{_{\text{KPM}}}$ and $\tau_{_{\text{RSA}}}$ is set to 1.0.

\begin{table}
   \small
   \begin{center}
      \begin{tabular}{p{1.8cm}<{\centering}p{1.8cm}<{\centering}p{1.5cm}<{\centering}p{1.5cm}<{\centering}}
         \toprule
         \multicolumn{2}{c}{Temperature}&
         \multicolumn{2}{c}{Evaluation Metric}\\
         $\tau_{_{\text{KPM}}}$ for KPM & $\tau_{_{\text{RSA}}}$ for RSA &mAP &top-1\\
         \midrule
         0.05&1.0&\textbf{89.2}&\textbf{91.1}\\
         0.5&1.0&87.8&88.8\\
         1.0&1.0&84.2&86.2\\
         0.05&0.5&89.0&89.6\\
         0.05&0.1&89.0&89.6\\
         \bottomrule
      \end{tabular}
   \end{center}
   \vspace{-5pt}
   \caption{mAP and top-1 accuracy by setting values for $\tau_{_{\text{KPM}}}$ and $\tau_{_{\text{RSA}}}$ for our proposed model on CUHK03~\cite{li2014deepreid}.}
   \label{tab:temp}
   \vspace{-15pt}
\end{table}

\section{Conclusion}
In this paper, we proposed an hourglass-like deep learning network Kronecker Product Matching, Soft Warping and Residual Self Attention for person re-identification. To match the correspondence feature between query image pair and further align the feature maps of query images,  our proposed method incorporates Kronecker Product Matching and Soft Warping in the training process which is an end-to-end manner.  Residual Self-Attention layer is also exploited for denoising the background noises. Furthermore, for capturing information from different scales we adopt hourglass structure in our model which could let model learn more robust and powerful feature.  The proposed approach outperforms state-of-the-art methods on three large person re-identification datasets. Extensive component analysis of our framework demonstrates the effectiveness of our overall framework and individual components.


{\small
\bibliographystyle{ieee}
\bibliography{egbib}
}

\end{document}